\documentclass[10pt,twocolumn,letterpaper]{article}

\usepackage{cvpr}              
\usepackage{multirow} 
\usepackage{longtable}
\usepackage{xtab}

\usepackage{colortbl}
\usepackage{array}
\usepackage{makecell}

\definecolor{cvprblue}{rgb}{0.21,0.49,0.74}
\usepackage[pagebackref,breaklinks,colorlinks,allcolors=cvprblue]{hyperref}

\def\confName{CVPR}
\def\confYear{2026}

\title{From Panel to Pixel: Zoom-In Vision-Language Pretraining from Biomedical Scientific Literature}

\author{
Kun Yuan$^{1,2,6,7}$ \quad
Min Woo Sun$^{3}$ \quad
Zhen Chen$^{4,9}$ \quad
Alejandro Lozano$^{3}$ \quad
Xiangteng He$^{5,8}$ \\ 
Shi Li$^{1,7}$ \quad
Nassir Navab$^{2}$ \quad
Xiaoxiao Sun$^{3}$ \quad
Nicolas Padoy$^{1,7}$ \quad
Serena Yeung-Levy$^{3}$
\and
$^{1}$University of Strasbourg, CNRS, INSERM, ICube, UMR7357, Strasbourg, France \\
$^{2}$Technical University of Munich \quad
$^{3}$Stanford University\quad
$^{4}$DSAI, The Hong Kong Polytechnic University  \\
$^{5}$University of British Columbia \quad
$^{6}$Munich Center for Machine Learning \\
$^{7}$IHU Strasbourg, Strasbourg \quad
$^{8}$Vector Institute for AI \quad
$^{9}$Yale University\\
{\tt\small kun.yuan@ext.ihu-strasbourg.eu}
}

\begin{document}
\maketitle
\begin{abstract}

There is a growing interest in developing strong biomedical vision-language models. A popular approach to achieve robust representations is to use web-scale scientific data. However, current biomedical vision-language pretraining typically compresses rich scientific figures and text into coarse figure-level pairs, discarding the fine-grained correspondences that clinicians actually rely on when zooming into local structures. To tackle this issue, we introduce Panel2Patch, a novel data pipeline that mines hierarchical structure from existing biomedical scientific literature, i.e., multi-panel, marker-heavy figures and their surrounding text, and converts them into multi-granular supervision. Given scientific figures and captions, Panel2Patch parses layouts, panels, and visual markers, then constructs hierarchical aligned vision-language pairs at the figure, panel, and patch levels, preserving local semantics instead of treating each figure as a single data sample. Built on this hierarchical corpus, we develop a granularity-aware pretraining strategy that unifies heterogeneous objectives from coarse didactic descriptions to fine region-focused phrases. By applying Panel2Patch to only a small set of the literature figures, we extract far more effective supervision than prior pipelines, enabling substantially better performance with less pretraining data.

\end{abstract}    
\section{Introduction}
\label{sec:intro}

\begin{figure}
    \centering
    \includegraphics[width=1\linewidth]{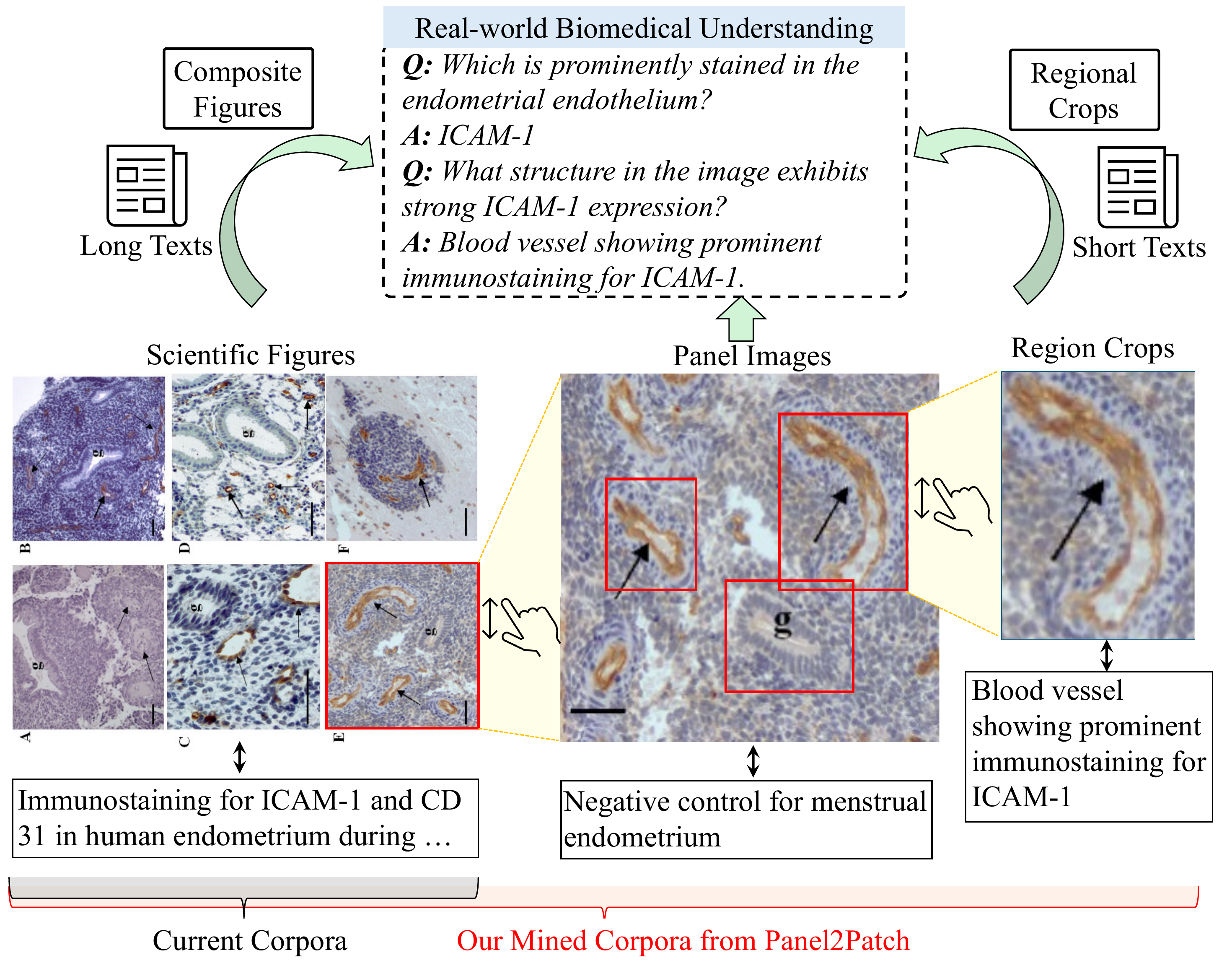}
    \caption{Our Panel2Patch pipeline generates
    additional fine-grained vision-language supervision signals and enhances the multi-modal representation through cross-level message passing.
    }
    \label{fig:fig1}
\end{figure}

Generalist vision--language foundation models are rapidly becoming central to biomedical research by learning unified representations across radiology~\cite{chen2024chexagent}, cellular microscopy~\cite{zhang2023biomedclip}, and surgical videos~\cite{yuan2025procedureawaresurgicalvideolanguagepretraining,yuan2025learningmultimodalrepresentationswatching} from large-scale image--text pairs. A widely adopted strategy for constructing supervision is inspired by how human experts are trained: exposure to textbooks and educational articles that provide rich multimodal content capturing expert knowledge, visual patterns, and their semantic associations, enabling strong generalization to tasks such as visual question answering~\cite{moor2023medflamingo} and image captioning~\cite{lee2025cxrllava}. However, these models often struggle when tasks require fine-grained discrimination between similar attributes, spatial configurations, or subtle semantic distinctions. This gap stems from the nature of scientific corpora, where figures are mostly multi-panel and captions provide high-level summaries, leading to coarse figure-level pairs for pretraining~\cite{sun2025no,lozano2025biomedicaopenbiomedicalimagecaption, luo2025vividmedvisionlanguagemodel} that favor thematic alignment but lack the localized grounding clinicians rely on when zooming into specific structures, as illustrated in Fig.~\ref{fig:fig1}.

\begin{table*}[t]
\centering
\scriptsize
\setlength{\tabcolsep}{6pt}
\renewcommand{\arraystretch}{1.15}
\caption{Comparison of vision-language data generation pipelines. Our Panel2Patch produces multi-panel, single-panel, and region-level correspondences, covering all granularities with automatic hierarchical captions and no additional annotation cost.
}
\begin{tabular}{ccccccccc}
\toprule
\textbf{Dataset} & \textbf{Multi-panel} & \textbf{Single-panel} & \textbf{Region} & \textbf{Captions} & \textbf{Grounding} & \textbf{Human Annotation} & \textbf{Cost} \\
\midrule
FG-CLIP~\cite{xie2025fgclipfinegrainedvisualtextual}  & $\times$ & $\checkmark$ & $\checkmark$  & CogVLM2 & Yolo-World & No & \textcolor{orange}{High} \\
GRIT~\cite{peng2023kosmos2} & $\times$ & $\checkmark$ & $\times$ & Annotated Captions & GLIP & No & \textcolor{orange}{High} \\
FineCLIP~\cite{wu2024fineclip} & $\times$ & $\checkmark$ & $\checkmark$ & BLIP2 & YOLO & Yes  & \textcolor{orange!70!black}{Moderate} \\
\hline
MEDTRINITY-25M~\cite{xie2025medtrinity25mlargescalemultimodaldataset} & $\times$ & $\checkmark$ & $\checkmark$  & GPT-4v + LLaVA\_MedCap & Expert Models \& GT & Yes & \textcolor{orange!70!black}{Moderate} \\
Biomedica~\cite{lozano2025biomedicaopenbiomedicalimagecaption} & $\checkmark$ & $\times$ & $\times$  & Literature Captions & N/A & Yes  & \textcolor{green!50!black}{Low} \\
Open-PMC-18M~\cite{baghbanzadeh2025open} & $\checkmark$ & $\checkmark$ & $\times$ & Literature Captions & YOLO & No  & \textcolor{green!50!black}{Low} \\
Ours & $\checkmark$ & $\checkmark$ & $\checkmark$ & Hierarchical Captions & QWen2.5-VL & No & \textcolor{green!50!black}{Low} \\
\bottomrule
\label{tab:dataset_comp}
\end{tabular}
\end{table*}

Existing vision-language data generation pipelines face a fundamental tradeoff between scalability and fine-grained supervision, as shown in Tab.~\ref{tab:dataset_comp}. Natural computer vision methods~\cite{li2022glip,xie2025fgclipfinegrainedvisualtextual,xie2025fgclip2,wu2024fineclip} achieve fine-grained region-level alignment through human annotations, pretrained detectors, or specialized captioners, but such annotation-intensive pipelines incur high computational costs and poorly generalize to biomedical domains due to domain gaps and imaging diversity. Conversely, scalable approaches that mine from biomedical scientific literature sacrifice granularity. They either treat multi-panel figures as single coarse instances~\cite{zhang2023biomedclip,lozano2025biomedicaopenbiomedicalimagecaption} or decompose them into panels~\cite{baghbanzadeh2025open} while still retaining figure-level captions that reference multiple elements, thus resulting in weak and coarse image-text alignment and missing the fine-grained region-level correspondences. Fundamentally, none of these methods produce hierarchical supervision spanning all three granularities, i.e., multi-panel figures, individual panels, and fine-grained regions, with corresponding hierarchical captions that capture both global context and localized semantics. This absence of multi-level, semantically grounded supervision limits current models' ability to learn fine-grained visual-textual correspondences in a scalable manner, hindering precise biomedical multimodal understanding.

To address these challenges, we propose a comprehensive approach that breaks the scalability-granularity tradeoff through two synergistic contributions. First, we introduce \textbf{Panel2Patch}, a data generation pipeline that exploits a key insight: \textit{scientific figures already contain hierarchical visual structure and explicit localization cues that can be automatically extracted}. Rather than requiring expensive annotations or specialized detectors, Panel2Patch leverages the inherent pedagogical design of scientific literature, i.e., multi-panel layouts, panel identifiers, and embedded visual markers (arrows, bounding boxes, zoom-in crops), to automatically mine supervision at three granularities: figure, panel, and patch/region. Using off-the-shelf vision-language models with \textit{set-of-marks} (SoM) prompting, it parses panel structures, decomposes captions, and localizes patches by detecting the visual markers authors use to highlight regions of interest. This yields hierarchical image-text pairs from global figure context to localized patch descriptions, achieving multi-level, fine-grained supervision at scale with minimal cost, as shown in Tab.~\ref{tab:dataset_comp}. Panel2Patch establishes a methodology for mining hierarchical supervision from any domain where visual documentation follows pedagogical conventions, offering a scalable alternative to manual annotation pipelines for biomedicine and other scientific fields to build their foundation models.

We also design a hierarchical zoom-in pretraining framework that uses these multi-granular pairs to train a single CLIP-style encoder whose panel-level embeddings serve as the primary representation for biomedical downstream tasks. The encoder is optimized jointly with auxiliary figure- and patch-level objectives and inter-level message passing, so that panel embeddings are explicitly refined by global figure context and fine-grained regional evidence, through top-down context propagation from figures to panels with global guidance and bottom-up evidence aggregation from patches to panels with localized cues. This hierarchical learning strategy produces a unified encoder where the figure and patch levels serve as contextual scaffolding that enhances the panel-level representations, while maintaining robust visual understanding across all granularities.

Together, these achieve data-efficient learning, outperforming prior biomedical VLP models trained on $\geq10\times$ larger corpora. Our contributions are threefold:
\begin{itemize}
\item We propose Panel2Patch, a scalable pipeline that automatically extracts hierarchical supervision (figure, panel, patch) from scientific literature's pedagogical structure.
\item We design a hierarchical zoom-in pretraining framework with inter-level message passing that enhances biomedical multimodal representations using multi-granularity context and fine-grained correspondences.
\item We achieve state-of-the-art performance on multiple external standard biomedical benchmarks, i.e., PatchCamelyon~\cite{Veeling2018-qh}, $\mu$-bench~\cite{lozano2024mu},  MedMNIST~\cite{yang2023medmnist}, LC25000~\cite{borkowski2019lung}, and Chexpert~\cite{irvin2019chexpert}, across different biomedical specialties, using 60\% less data than prior work, which paves a efficient path towards biomedical foundation model.
\end{itemize}

\section{Related Work}
\label{sec:related_work}

\subsection{Biomedical Vision-Language Pretraining}
Recent biomedical vision--language models mostly scale literature-derived image--text pairs. Early work such as PMC-CLIP \cite{lin2023pmcclip} trains on 1.6M figure--caption pairs, while BiomedCLIP \cite{zhang2023biomedclip} and BIOMEDICA \cite{lozano2025biomedicaopenbiomedicalimagecaption} expand to 15M and 24M pairs, respectively, improving domain-specific representations via larger corpora and richer metadata. However, these pipelines treat each scientific figure as a single image-level instance, even when figures are multi-panel with captions that mix global context and panel-specific details, leading to coarse supervision misaligned with how clinicians actually zoom into panels and local regions.

To obtain finer supervision, {MedTrinity-25M} \cite{xie2025medtrinity25mlargescalemultimodaldataset} augments biomedical corpora with expert-verified ROIs and GPT-4V/LLaVA-based captions, while FineCLIP~\cite{wu2024fineclip} and FG-CLIP~\cite{xie2025fgclipfinegrainedvisualtextual,xie2025fgclip2} build region--text datasets using detectors and task-specific captioners. Open-PMC-18M~\cite{baghbanzadeh2025open} segments figures into panels but reuses figure-level captions, so panel crops still carry entangled multi-panel descriptions. These works show that panel- and region-level signals are beneficial, but they either depend on manual labels, specialized detectors, or expensive captioners, and still do not capture the full figure--panel--region hierarchy.

Our work is complementary: instead of collecting more data or adding human labels, we mine richer supervision from existing biomedical corpora. Panel2Patch explicitly exploits the pedagogical structure of scientific figures---multi-panel layouts, panel identifiers, arrows, zoom-in crops, and other visual markers---to automatically construct aligned image--text pairs at figure, panel, and region levels. Unlike prior biomedical VLP pipelines that assume a single granularity, we recover hierarchical supervision directly from the literature and show that training on this structure yields stronger and more data-efficient foundation models.

\subsection{Fine-Grained and Hierarchical Alignment}

In the general domain, fine-grained vision-language alignment has been achieved either by (i) region-based methods that rely on object detectors and region-phrase contrastive learning~\cite{li2022groundedlanguageimagepretraining,zhong2021regionclipregionbasedlanguageimagepretraining,xie2025fgclip2,xie2025fgclipfinegrainedvisualtextual,wu2024fineclip, li2022finegrainedsemanticallyalignedvisionlanguage, yao2021filip, jin2023grillgroundedvisionlanguagepretraining, li2025unbiasedregionlanguagealignmentopenvocabulary, minderer2024scalingopenvocabularyobjectdetection}, or (ii) dense feature methods that compute token-level similarities between visual and textual embeddings to derive soft correspondence maps~\cite{rao2022densecliplanguageguideddenseprediction,yao2021filip, zhang2024longclipunlockinglongtextcapability, yang2023alipadaptivelanguageimagepretraining, wu2023coraadaptingclipopenvocabulary, luo2023segclippatchaggregationlearnable, wu2023aligningbagregionsopenvocabulary}. While these approaches are effective on natural images, they are typically trained and evaluated in settings with detector backbones and large-scale region-level annotations, and assume scenes composed of continuously varying objects in a single frame. In contrast, biomedical scientific figures behave more like structured documents: they consist of discrete, alphabetically labeled panels and visually explicit markers such as arrows, brackets, and zoom-in insets, and there is no analogous large-scale, detector-ready region-annotation corpus in this domain.

\begin{figure*}[!ht]
    \centering
    \includegraphics[width=0.8\linewidth]{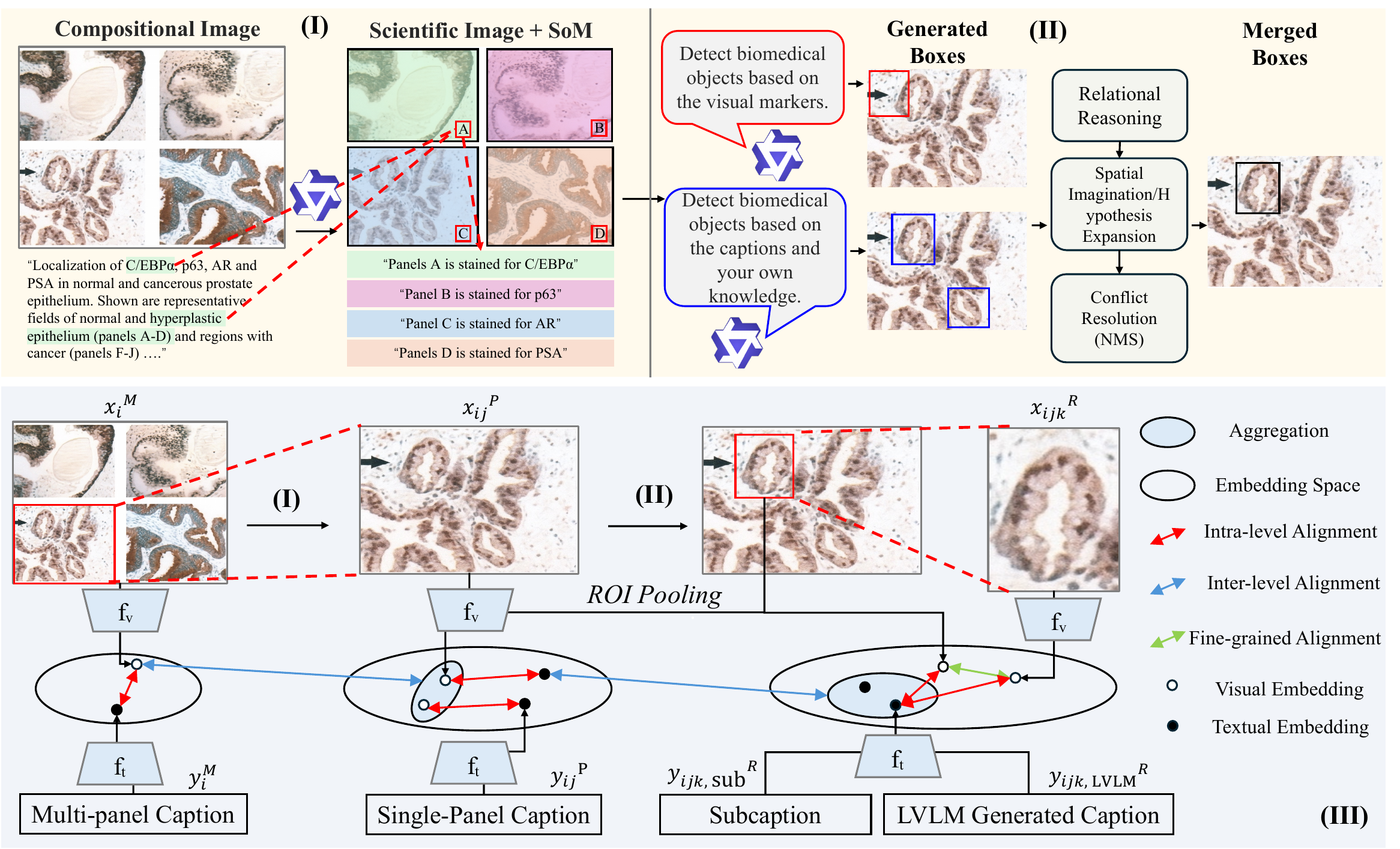}
    \caption{Overview of our Panel2Patch pipeline (I, II) and hierarchical pretraining (III). (I) Multi-panel figures are split into single panels and associated with their captions. 
    (II) Complementary LVLM-based agents detect accurate biomedical regions of interest. 
    (III) Hierarchical pretraining learns from multi-level vision–language correspondences and enhances embedding space across panel-, patch-, and region-level representations.}
    \label{fig:fig2}
\end{figure*}

A complementary line of work explores hierarchical alignment by building multi-scale embeddings over images or videos~\cite{gao2022pyramidcliphierarchicalfeaturealignment,ye2023hiervlhierarchicalvideotext, sun2023alphaclipclipmodelfocusing, ashutosh2023hiervllearninghierarchicalvideolanguage, ma2022xclipendtoendmultigrainedcontrastive}, for example across different spatial resolutions or clip/scene/video levels. These methods implicitly infer hierarchy from visual features and are tailored to natural scenes and videos, rather than to multi-panel scientific figures with explicitly encoded structure. As a result, they cannot directly address the figure-panel-region layout that dominates biomedical literature figures. Therefore, we focus on comparisons with biomedical VLP models trained on literature-derived figure-caption corpora, where Panel2Patch is designed to be a drop-in improvement by upgrading the supervision from coarse figure-level pairs to a structured figure-panel-region hierarchy. 

\section{Methodology}

We aim to train a single CLIP-style vision-language encoder that serves as a drop-in panel-level backbone for standard biomedical tasks, while also supporting multi-panel and region-level retrieval/grounding. To enable this, \textbf{Panel2Patch} first converts biomedical figures and their surrounding text into aligned image-text pairs at three granularities, i.e., full figures, single panels, and marker-indicated regions. On top of this corpus, we then apply a \textbf{hierarchical zoom-in pretraining} that maps all three granularities into one shared embedding space, where panel-level embedding space is explicitly optimized by auxiliary figure and region-level image-text pairs via inter-level message passing.

\subsection{Panel2Patch Data Generation Pipeline}

Given a scientific figure and its paired caption, \textbf{Panel2Patch} produces hierarchical supervision at three levels: figure, panel, and patch, as shown in Fig.~\ref{fig:fig2}. 
While prior works in natural image domain benefit from domain-specific captioning and grounding models trained with dense region-text annotations, the biomedical domain lacks such curated resources. We tackle this gap by exploiting our central insight that \textit{scientific figures already contain hierarchical visual structure and explicit localization cues that can be automatically extracted}, enabling us to approximate region-level grounding without explicit region-text labels. Panel2Patch leverages off-the-shelf LVLMs and their ability to localize visual markers, e.g., set-of-markers (SOM) and arrows, to turn existing biomedical scientific figures into rich multi-level supervision with high-quality fine-grained bounding box-text pairs.

\subsubsection{SoM-Guided Panel Decomposition}

\textbf{Panel proposal with SoM-style prompting.} As illustrated in Fig.~\ref{fig:fig2} (I), we treat the visual markers already overlaid on biomedical figures, e.g., ``A'', ``I'', as implicit Set-of-Mark (SoM) cues~\cite{yang2024set}. For each compositional figure, we prompt an LVLM to (i) propose rectangular regions corresponding to visually coherent panels and (ii) read the nearby overlaid label and use it as the panel identifier (``A'', ``I'', \dots). To improve robustness, we repeat this query over multiple random scales and crops, aggregate the predictions, and apply NMS over boxes sharing the same identifier. This yields a compact set of panel bounding boxes, which we use to crop the original figure into single-panel images for subsequent text association and patch mining. We also train an image classifier to detect and discard non-photographic plots, e.g., bar charts, to extract biomedical meaningful supervisions.

\textbf{Panel-aware text association.} Given the detected panels and their identifiers, we assign text to each panel in an identifier-centric manner. We first decompose the composite caption and nearby narrative into sentences and short clauses, using occurrences of panel identifiers as anchors. The LVLM is then prompted to segment the caption into minimal semantic units and assign each fragment to the panel(s) whose SoM-style marks share the same identifier. This identifier-driven assignment avoids most biomedical vision-language reasoning, making the mapping tractable even for a general-purpose LVLM without biomedical domain-specific finetuning. Finally, we ask the LVLM to generate a short additional description for each panel, conditioned on both the panel crop and its assigned fragments, leveraging its biomedical knowledge to enrich panel text and strengthen the pretraining signal~\cite{wang2025advancing,fan2023improving}. We manually assess the quality of this supervision by inspecting 2{,}000 images and find that roughly $\sim 80\%$ of compositional figures are correctly decomposed into single-panel images.

\subsubsection{Marker-Guided Region Mining and Captioning}
\label{sec:region_proposal}

When mining fine-grained region-text pairs, existing pipelines typically rely on a pretrained detector to propose regions and then use either a trained captioning model or a large VLM to generate descriptions, which depends on large-scale annotated detection/caption data and remains vulnerable to hallucinated regions and spurious text.

Biomedical figures, however, frequently use arrows, brackets, and color overlays to highlight salient structures, and we exploit these cues to mine region-level supervision with a marker-anchored fusion of two LVLM-derived proposals, as illustrated in Fig.~\ref{fig:fig2} (II). We first detect \textbf{marker} boxes that tightly frame visual markers, e.g., arrows, asterisks, and \textbf{caption} boxes proposed from the corresponding panel captions. Caption-based boxes are retained only if their centers lie within a normalized distance $\le \tau$ of any marker center, ensuring that textual proposals remain anchored to explicit visual cues, while markers without nearby caption proposals are expanded around their centers to approximate the extent of the referenced object. We then take the union of kept caption boxes and inflated marker boxes and apply IoU-based NMS, yielding a compact set of regions that stay spatially consistent with marker cues while suppressing spurious boxes.

For each selected region, we attach text via two complementary LVLM pathways. Given the panel and its caption, the LVLM decomposes long sentences into clauses and grounds fragments to marker-indicated areas, using keywords such as ``arrow'' or ``asterisk'' as anchors to route relevant phrases to the corresponding boxes. In parallel, we query the LVLM directly on each cropped region to generate a short description that emphasizes local visual details, e.g., morphology, staining intensity, fine structure. We keep both the grounded caption fragments and the patch-level descriptions for pretraining, providing region-level supervision that is visually faithful and aligned with the figure narrative. As shown in Fig.~\ref{fig:fig2}, fusing marker- and caption-based proposals in this way reduces LVLM hallucinations: caption boxes that drift to unrelated areas are filtered by the marker-proximity gate, while marker boxes are expanded only locally rather than allowing the model to invent new regions. The resulting regional image-text pairs form a high-precision corpus for fine-grained vision-language alignment.

\subsection{Zoomed In Vision-Language Pretraining Method for Hierarchical Attributed Scientific Figures}
\label{sec:pretraining}

As described above, our Panel2Patch converts biomedical scientific figures into aligned image-text pairs at three levels: multi-panel figures, single panels, and regions. To exploit this structure during learning, we adopt a zoomed-in pretraining framework that models image-text correspondences at these three granularities, i.e., multi-panel (\(M\)), single-panel (\(P\)), and regional (\(R\)). 

\subsection{Hierarchical Image-text Pairs}
Our data comprises three granularities of supervision. At the coarsest level, each multi-panel figure \(i\) provides an image-text pair \((x_i^{M}, y_i^{M})\), where \(x_i^{M}\) is the full figure and \(y_i^{M}\) its global caption (the superscript $M$ denotes multi-panel level). At the intermediate level, the figure contains panels indexed by \(j \in \mathcal{P}(i)\) (panel $j$ belongs to the set of panels in figure $i$), each with an image \(x_{ij}^{P}\) and a localized caption \(y_{ij}^{P}\) (the superscript $P$ denotes panel level). At the finest level, our marker-guided mining (Sec.~\ref{sec:region_proposal}) identifies regions indexed by \(k \in \mathcal{R}(i,j)\) (region $k$ belongs to the set of regions in panel $j$ of figure $i$), yielding regional crops \(x_{ijk}^{R}\) and two complementary texts: an LVLM-generated patch caption \(y_{ijk,\mathrm{LVLM}}^{R}\) and a marker-grounded sub-caption \(y_{ijk,\mathrm{sub}}^{R}\) (the superscript $R$ denotes region level). During pretraining, one of these two texts is randomly sampled per region. Note that not all biomedical scientific figures are multi-panel, and not all single-panel images contain valid region annotations, therefore, the number of available pairs at the \(M\), \(P\), and \(R\) levels differs.

\begin{table*}[t]
\centering

\renewcommand{\arraystretch}{1.0} 
\caption{Our model is remarkably better than prior biomedical VLP models on single-panel short-context retrieval (Panel A), due to the cross-level refinement from multi-panel and region-level image-text pairs. We also achieve a slightly better performance (Panel B) due to our extracted fine-grained supervision signals from Panel2Patch.}
\label{tab:three_panels_retrieval}
\begin{tabular}{l l ccc ccc}
\toprule
\textbf{Model} & \textbf{Setting} & \multicolumn{3}{c}{\textbf{I2T}} & \multicolumn{3}{c}{\textbf{T2I}} \\
 &  & R@1 (\%) & R@5 (\%) & R@10 (\%) & R@1 (\%) & R@5 (\%) & R@10 (\%) \\
\midrule
\multicolumn{8}{l}{\textit{Panel A: Single-panel short-context retrieval}} \\
OPEN\_CLIP~\cite{radford2021learning}        & Short-context     & 14.78 & 35.95 & 45.26 & 18.79 & 37.09 & 48.69 \\ 
PMC-CLIP~\cite{lin2023pmcclip}               & Short-context     & 0.05  & 0.23  & 0.46  & 0.05  & 0.28  & 0.46  \\
MedSigLIP~\cite{sellergren2025medgemma}      & Short-context     & 10.62 & 23.69 & 32.03 & 9.31  & 22.06 & 31.05 \\
BioMedCLIP~\cite{zhang2023biomedclip}        & Short-context     & 33.66 & \underline{65.36} & \underline{74.84} & 30.07 & 60.95 & \underline{73.20} \\
BMC-CLIP~\cite{lozano2025biomedicaopenbiomedicalimagecaption} & Short-context & \underline{34.15} & 64.71 & 73.53 & \underline{32.03} & \underline{62.75} & 73.86 \\
Ours                                      & Short-context     & \textbf{36.60} & \textbf{68.14} & \textbf{79.90} & \textbf{38.24} & \textbf{68.14} & \textbf{80.88} \\
\midrule
\multicolumn{8}{l}{\textit{Panel B: Bounding-box $\leftrightarrow$ Text retrieval}} \\
OPEN\_CLIP~\cite{radford2021learning}        & BBox$\leftrightarrow$Text & 3.74  & 10.30 & 15.25 & 5.64  & 12.38 & 16.77 \\
PMC-CLIP~\cite{lin2023pmcclip}               & BBox$\leftrightarrow$Text & 0.16  & 0.98  & 1.63  & 0.16  & 0.82  & 1.63  \\
MedSigLIP~\cite{sellergren2025medgemma}      & BBox$\leftrightarrow$Text & 2.26  & 5.50  & 8.09  & 1.43  & 4.53  & 6.52  \\
BioMedCLIP~\cite{zhang2023biomedclip}        & BBox$\leftrightarrow$Text & 5.59  & 14.42 & 20.84 & 6.38  & 16.91 & 22.74 \\
BMC-CLIP~\cite{lozano2025biomedicaopenbiomedicalimagecaption} & BBox$\leftrightarrow$Text & \underline{8.04}  & \textbf{20.98} & \textbf{27.82} & \underline{9.29}  & \underline{21.44} & \underline{28.42} \\
Ours                                      & BBox$\leftrightarrow$Text & \textbf{8.64}  & \underline{20.24} & \underline{27.73} & \textbf{9.38}  & \textbf{22.37} & \textbf{30.50} \\
\bottomrule
\end{tabular}
\end{table*}

\subsubsection{Hierarchical Embedding Spaces}

We design a pretraining framework tailored to these hierarchical {multi-panel}, {single-panel}, and {regional} image-text pairs. We use a shared image encoder $f_v$ and text encoder $f_t$ to map all inputs to $d$-dimensional embeddings, regardless of granularity. Specifically, for each figure $i$, panel $j$, and region $k$, we compute:
\[
\begin{aligned}
\mathbf{v}_i^{M} &= f_v(x_i^{M}),\quad  \mathbf{t}_i^{M} = f_t(y_i^{M}),\\
\mathbf{v}_{ij}^{P} &= f_v(x_{ij}^{P}),\quad  \mathbf{t}_{ij}^{P} = f_t(y_{ij}^{P}),\\
\mathbf{v}_{ijk}^{R} &= f_v(x_{ijk}^{R}),\quad  \mathbf{t}_{ijk}^{R} = f_t(y_{ijk}^{R}),
\end{aligned}
\]
where $\mathbf{v}$ denotes visual embeddings and $\mathbf{t}$ denotes text embeddings at each level. We then construct a unified hierarchical embedding space by jointly enforcing {intra-level}, {inter-level}, and {fine-grained} alignment.

\textbf{Intra-level Alignment.} Within each granularity, we align images and their corresponding text descriptions using standard contrastive learning. Specifically, at the multi-panel level ($M$), we form positive pairs from figures and their captions; at the panel level ($P$), from panels and their localized captions; and at the region level ($R$), from patches and their fine-grained descriptions. These positive sets are denoted as:
\[
\mathcal{P}_\mathrm{intra}^{M},\quad \mathcal{P}_\mathrm{intra}^{P},\quad \mathcal{P}_\mathrm{intra}^{R},
\]
which are optimized with the corresponding CLIP losses
$\mathcal{L}_\mathrm{intra}^{M}$, $\mathcal{L}_\mathrm{intra}^{P}$, and $\mathcal{L}_\mathrm{intra}^{R}$, as shown in Fig.~\ref{fig:fig2}.
This captures intra-level correspondences while respecting the distinct distributions of $M/P/R$.

\textbf{Fine-grained Alignment.}
Biomedical image understanding requires the region-level grounding and understanding of localized regions, therefore, we impose explicit fine-grained alignment between regions and their textual descriptions, where region descriptions are paired with corresponding image regions. Additionally, we apply ROI pooling on the panel's feature map to extract region-specific feature crops, as shown in Fig.~\ref{fig:fig2}, enforcing consistency between pixel-level and feature-level representations. This fine-grained alignment not only enables region-level tasks such as bounding box retrieval and grounding, but also refines the main panel-level embedding space through bottom-up message passing, ensuring that panel representations are grounded in localized visual evidence and corresponding textual semantics.

\textbf{Inter-level Message Passing.} Biomedical scientific figures encode semantics across hierarchies, i.e., a multi-panel figure encodes global context for its single-panel children, while each region conveys localized evidence for its parent panel. To exploit this structure, we refine each granularity with semantics from the others by passing messages across levels. Specifically, we align coarse and fine embeddings through aggregated summaries that propagate contextual information upward (from fine to coarse) and gradients downward (from coarse to fine). We first aggregate fine-level embeddings into coarse summaries using average pooling:
\[
\begin{aligned}
\bar{\mathbf{v}}_i^{P} &= \mathrm{Agg}\!\big(\{\mathbf{v}_{ij}^{P}\}_{j\in\mathcal{P}(i)}\big),\\
\bar{\mathbf{t}}_i^{P} &= \mathrm{Agg}\!\big(\{\mathbf{t}_{ij}^{P}\}_{j\in\mathcal{P}(i)}\big),\\
\bar{\mathbf{v}}_{ij}^{R} &= \mathrm{Agg}\!\big(\{\mathbf{v}_{ijk}^{R}\}_{k\in\mathcal{R}(i,j)}\big).
\end{aligned}
\]
Here, $\bar{\mathbf{v}}_i^{P}$ is the aggregated visual embedding of all panels in figure $i$, $\bar{\mathbf{t}}_i^{P}$ is the aggregated text embedding of all panel captions in figure $i$, and $\bar{\mathbf{v}}_{ij}^{R}$ is the aggregated visual embedding of all regions in panel $j$. For multi-panel $\leftrightarrow$ single-panel consistency, we then align the full figure embedding with its aggregated panel embeddings by defining inter-level positive pairs as:
\[
\mathcal{P}_\mathrm{inter}^{M\leftrightarrow P}
=\{(\mathbf{v}_i^{M},\bar{\mathbf{v}}_i^{P}),\;(\mathbf{t}_i^{M},\bar{\mathbf{t}}_i^{P})\}_i,
\]
This set contains pairs matching the figure's visual embedding $\mathbf{v}_i^{M}$ with the aggregated panel visual embeddings $\bar{\mathbf{v}}_i^{P}$, and similarly for text embeddings. We optimize these with a CLIP loss:
\[
\mathcal{L}_\mathrm{inter}^{M\leftrightarrow P}
= \mathcal{L}_{\mathrm{CLIP}}(\{\mathbf{v}_i^{M}\},\{\bar{\mathbf{v}}_i^{P}\})
+ \mathcal{L}_{\mathrm{CLIP}}(\{\mathbf{t}_i^{M}\},\{\bar{\mathbf{t}}_i^{P}\}).
\]
This ensures the figure embedding is consistent with what its constituent panels collectively encode. Analogously, we enforce single-panel $\leftrightarrow$ regional coupling to ensure each panel embedding summarizes its regions:
\[
\begin{aligned}
\mathcal{P}_\mathrm{inter}^{P\leftrightarrow R}
&= \{\,(\mathbf{v}_{ij}^{P},\bar{\mathbf{v}}_{ij}^{R}),\;(\mathbf{t}_{ij}^{P},\bar{\mathbf{t}}_{ij}^{R})\,\}_{i,j},\\
\mathcal{L}_\mathrm{inter}^{P\leftrightarrow R}
&= \mathcal{L}_{\mathrm{CLIP}}\big(\{\mathbf{v}_{ij}^{P}\},\{\bar{\mathbf{v}}_{ij}^{R}\}\big)
 + \mathcal{L}_{\mathrm{CLIP}}\big(\{\mathbf{t}_{ij}^{P}\},\{\bar{\mathbf{t}}_{ij}^{R}\}\big),
\end{aligned}
\]
\noindent where $\bar{\mathbf{t}}_{ij}^{R}=\mathrm{Agg}\!\big(\{\mathbf{t}_{ijk}^{R}\}_{k\in\mathcal{R}(i,j)}\big)$ is the aggregated text embedding of all regions in panel $j$. Together, these inter-level losses implement top-down context propagation and bottom-up evidence aggregation, yielding bidirectional message passing across $M\!\leftrightarrow\!P\!\leftrightarrow\!R$. This design benefits the panel-level embedding space, which serves as the primarily embedding space for biomedical downstream tasks: $\mathcal{L}_\mathrm{inter}^{M\leftrightarrow P}$ injects global context from figures, while $\mathcal{L}_\mathrm{inter}^{P\leftrightarrow R}$ enriches panels with fine-grained regional details.

\subsubsection{Alternating Training}
To prevent catastrophic forgetting across levels and to handle the strong data imbalance between figures, panels, and regions, we adopt a coarse-to-fine alternating schedule that activates one level per step (M$\rightarrow$P$\rightarrow$R). Cycling supervision ensures that recently updated parameters are revisited from other granularities before they drift, preserving previously learned semantics. We train the model in an alternating way, with a few batches of fine-grained regions are followed by single-panel batches, then by multi-panel batches. This alternating training avoids overfitting to single-level of data, and helps the model learn balanced representations across all granularities.

\section{Experiments}

\subsection{Implementation Details}
We use a ViT-L/14 vision encoder initialized from BMC-CLIP~\cite{lozano2025biomedicaopenbiomedicalimagecaption} and freeze the entire text encoder as well as the early layers of the vision tower, updating only the last five transformer blocks of the image encoder. The model is trained for 20 epochs with AdamW (weight decay $0.05$, $\beta_1{=}0.9$, $\beta_2{=}0.95$), a cosine learning rate schedule with $1{,}000$ warmup steps, and a base learning rate of $1\mathrm{e}{-5}$. Training is distributed over 8 GPUs with a per-GPU batch size of 40. We uniformly sample approximately 350k raw figures from Biomedica~\cite{lozano2025biomedicaopenbiomedicalimagecaption} and process them with the open-source LVLM Qwen2.5-VL-72B using SoM prompting to generate about 1M single-panel images. Panel parsing takes roughly 5 days on 8$\times$H100 GPUs ($\sim$960 GPU-hours), and our marker-guided region mining and caption generation pipeline on the same hardware requires an additional $\sim$10 days ($\sim$1,920 GPU-hours), for a total of about 2,900 GPU-hours. All stages can be straightforwardly parallelized across more GPUs to reduce wall-clock time. Further data statistics and implementation details are provided in the supplementary material.

\vspace{-1mm}

\subsection{Panel2Patch Enables Multi-Granularity Biomedical Vision--Language Tasks}

Panel2Patch provides a data-centric alternative to prior biomedical vision-language pipelines by automatically mining supervision at multiple levels, i.e., figures, panels, and regions, without additional human annotation. As summarized in Tab.~\ref{tab:dataset_comp}, existing datasets such as MEDTRINITY-25M~\cite{xie2025medtrinity25mlargescalemultimodaldataset} and FineCLIP~\cite{wu2024fineclip} rely on costly GPT-based captioning or manually verified grounding, while GRIT~\cite{peng2023kosmos2} and FG-CLIP~\cite{xie2025fgclipfinegrainedvisualtextual} depend on task-specific detectors and domain-specific tuning. In contrast, Panel2Patch uses a general LVLM with task-oriented prompts and figure-intrinsic cues, e.g., visual markers, to derive aligned supervision across figure-panel-region hierarchy, reducing annotation cost while preserving structured alignment at  3 levels.

We evaluate the resulting model on tasks that mirror this hierarchy. At the panel level, we consider single-panel retrieval and zero-shot classification, where the model needs to align concise textual descriptions with image-centric panels. At the region level, we evaluate bounding box$\leftrightarrow$text retrieval and phrase grounding, which require precise localization of marker-indicated structures. As shown in Tab.~\ref{tab:three_panels_retrieval}, Panel2Patch consistently improves over prior biomedical vision-language models, not only in coarse panel-level metrics but also in fine-grained region alignment. In practice, our model can support workflows that range from browsing panels by textual query to zooming into localized findings, matching how biomedical experts naturally read and interpret biomedical images.

\subsection{Panel2Patch Builds Data-Efficient Biomedical Vision-Language Foundation Model}

Beyond retrieval, we evaluate zero-shot classification across six biomedical specialties, as shown in Tab.~\ref{tab:model_vs_group_avg}. Using only a subset of prior biomedical data (400K pairs vs.\ $\sim$6M), Panel2Patch attains state-of-the-art performance, surpassing models trained on far larger corpora. This demonstrates that superior supervision quality can outweigh raw scale. Also, by mining hierarchical signals at the figure, panel, and region levels and enforcing fine-grained grounding, our Panel2Patch data pipeline supports versatile supervision signals that transfer across specialties. These findings demonstrate a practical pathway for building biomedical foundation models, by mining richer supervision from existing large-scale pretraining corpora, without changing the learning objective or the architecture, we can achieve stronger performance with the same data.

\begin{table*}[ht]
\centering
\small
\setlength{\tabcolsep}{5pt}
\renewcommand{\arraystretch}{1.2}
\begin{tabular}{l>{\columncolor{gray!15}}cc ccccc>{\columncolor{gray!15}}c}
\hline
\textbf{Model} & \textbf{Data Size} & \textbf{Biology} & \textbf{Dermatology} & \textbf{Microscopy} & \textbf{Ophthalmology} & \textbf{Pathology} & \textbf{Radiology} & \textbf{Avg.} \\
\hline
PMC-CLIP~\cite{lin2023pmcclip} & 80K & 7.75 & 12.59 & 10.91 & 23.26 & 19.11 & 38.64 & 18.71 \\
MedSigLIP~\cite{sellergren2025medgemma} & 33M & 32.09 & 20.13 & 34.90 & 38.23 & 40.17 & 44.76 & 35.38 \\
BiomedCLIP~\cite{zhang2023biomedclip} & 15M & 34.07 & 36.01 & 49.71 & 37.36 & 38.40 & 56.05 & 41.93 \\
BMC-CLIP~\cite{lozano2025biomedicaopenbiomedicalimagecaption} & 24M & 34.08 & \textbf{65.81} & \underline{50.09} & 36.74 & 41.21 & 59.15 & 47.85 \\
BMC-LongCLIP~\cite{sun2025no} & 1M & \textbf{40.82} & 40.69 & 46.04 & \textbf{59.80} & 42.28 & \underline{59.42} & \underline{48.18} \\

Ours & 400K & \underline{38.53} & \underline{65.19} & \textbf{54.00} & \underline{35.78} & \textbf{42.91} & \textbf{65.07} & \textbf{50.25} \\

\hline
\end{tabular}
\caption{Zero-shot classification performance comparison across five biomedical specialties. We evaluate models on external biomedical benchmarks and group their performance by specialty. These specialties cover a broad spectrum of biomedical imaging modalities and clinical tasks, enabling a comprehensive assessment of domain-wise generalization. Trained on a substantially smaller biomedical corpus, our model attains even superior performance, highlighting its data efficiency. Best result in bold and second best with underline.}

\label{tab:model_vs_group_avg}
\end{table*}


\begin{figure}
    \centering
    \includegraphics[width=0.9\linewidth]{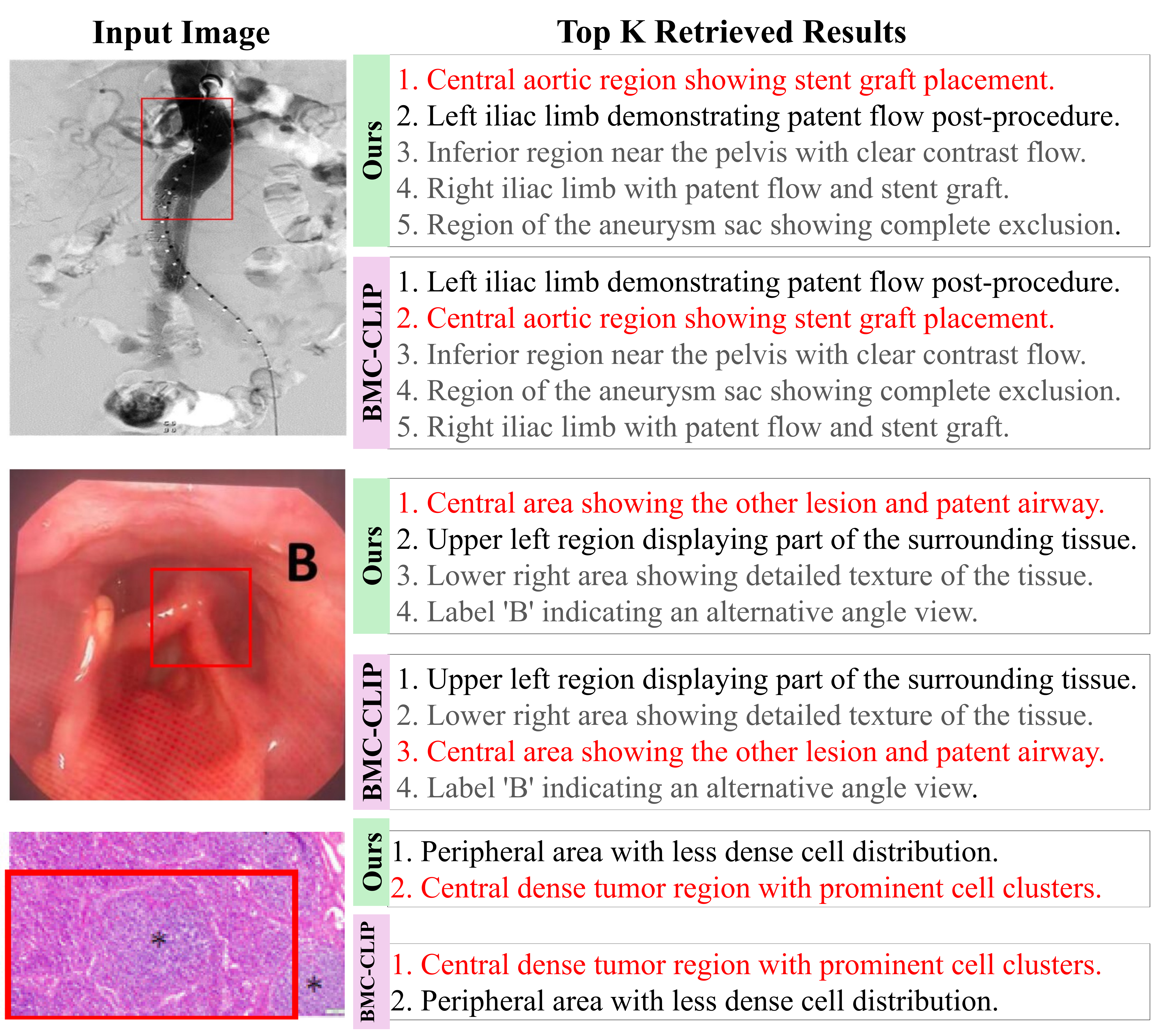}
    \caption{Qualitative examples of fine-grained retrieval using bounding box-cropped images and region-level texts.}
    \label{fig:fine-grained}
\end{figure}

\subsection{Panel2Patch Mines Fine-grained Patterns}

We further probe region-level representations by visualizing top-ranked box$\leftrightarrow$text matches and ROI-pooled alignments, as shown in Fig.~\ref{fig:fine-grained}. The model reliably localizes small visual entities and retrieves the corresponding phrases, yielding precise grounding of surgical instruments, tissue structures, and other localized biomedical findings. Beyond a single domain, Panel2Patch performs region-level understanding across radiology, microscopy, and cellular imaging without any task-specific tuning. Compared to prior state-of-the-art models that primarily learn coarse image-level correspondences, our model explicitly captures localized image-text alignment, making it directly useful for downstream tasks that require fine-grained grounding, such as grounded VQA and phrase-to-region retrieval.

\subsection{Ablation Study}
\begin{table}[t]
\centering
\caption{Ablation study on different level of data.  We use the same testing benchmarks as in Tab.~\ref{tab:three_panels_retrieval}}
\label{tab:ablation}
\resizebox{0.45\textwidth}{!}{%
\begin{tabular}{l cc cc}
\toprule
\textbf{Data} & \multicolumn{2}{c}{\textbf{I2T}} & \multicolumn{2}{c}{\textbf{T2I}} \\
&  R@5 (\%) & R@10 (\%) & R@5 (\%) & R@10 (\%) \\
\midrule
\multicolumn{5}{l}{\textit{Panel A: Single-panel short-context retrieval.}} \\
CLIP~\cite{radford2021learning}           & 35.95 & 45.26 & 37.09 & 48.69 \\ 
Single-panel  & \underline{66.67} & \underline{77.61} & \underline{63.73} & \underline{76.96} \\ 
Region        & 61.60 & 73.37 & 61.11 & 72.55 \\
Panel2Patch   & \textbf{68.14} & \textbf{79.90} & \textbf{68.14} & \textbf{80.88} \\
\midrule
\multicolumn{5}{l}{\textit{Panel B: Bounding-box $\leftrightarrow$ Text retrieval}} \\
N/A           & 10.30 & 15.25 & 12.38 & 16.77 \\
Single-panel  & 17.65 & 23.66 & 19.50 & 26.52 \\
Region        & \underline{18.90} & \underline{25.65} & \underline{20.52} & \underline{27.08} \\
Panel2Patch   & \textbf{20.24} & \textbf{27.73} & \textbf{22.37} & \textbf{30.50} \\
\bottomrule
\end{tabular}%
}
\end{table}

\paragraph{Effect of Hierarchical Embedding Space.}
We probe how each supervision level contributes by comparing models trained with only region data, only single-panel data, and our full Panel2Patch setup. Removing region-level supervision hurts fine-grained grounding. Specifically, compared to our full model, a single-panel–only variant shows a clear drop in box$\leftrightarrow$text retrieval, as shown in Tab.~\ref{tab:ablation}, indicating that zoomed-in supervision is necessary to align marker-indicated structures. Conversely, removing single-panel data breaks local understanding. Region-only training yields lower short-context panel retrieval than training with only single-panel data, showing that region supervision alone cannot recover panel-level semantics.

\begin{table}[t]
\centering
\caption{
Cross-modal retrieval performance when training on single-panel, multi-panel, or our alternating training regimen. Our alternating training maintains performance on the multi-panel level while achieving better performance at the finer single-panel scale. Results are computed on 100K image-text pairs sampled from our full pretraining data.
}
\label{tab:alt_single_multi}
\small
\setlength{\tabcolsep}{3pt}
\begin{tabular}{lcccccc}
\toprule
\textbf{Train $\rightarrow$ Test} &
\multicolumn{3}{c}{\textbf{I2T}} &
\multicolumn{3}{c}{\textbf{T2I}} \\
& \makecell{R@1\\(\%)} & \makecell{R@5\\(\%)} & \makecell{R@10\\(\%)} & \makecell{R@1\\(\%)} & \makecell{R@5\\(\%)} & \makecell{R@10\\(\%)} \\
\midrule
Multi $\rightarrow$ Single      & 1.95 & 5.46 & 8.14 & 1.50 & 4.82 & 7.71 \\
Multi $\rightarrow$ Multi       & 8.68 & 21.54 & 29.35 & 8.04 & 20.76 & 28.35 \\
\midrule
Single $\rightarrow$ Single     & {4.38} & {11.11} & {15.36} &
                                  {4.07} & {10.52} & {15.51} \\
Single $\rightarrow$ Multi      & 2.77 & 8.20 & 12.37 & 3.10 & 9.85 & 14.87 \\
\midrule
Ours $\rightarrow$ Single       & 4.34 & 10.72 & 14.79 & 3.84 & 10.53 & 15.19 \\
Ours $\rightarrow$ Multi        & 8.74 & 22.07 & 29.59 & 8.93 & 22.15 & 29.94 \\
\bottomrule
\end{tabular}
\end{table}

\textbf{Effect of Alternating Training.}
As shown in Tab.~\ref{tab:alt_single_multi} We test whether our alternating between figure, panel, and region supervision truly helps balance learning across levels. We trained three versions: one using only multi-panel data, one using only single-panel data, and one using our alternating schedule. As shown in Tab.~\ref{tab:alt_single_multi}, the single-level models quickly overfit and the multi-only model worked well on multi-panel retrieval but failed at fine scales, while the single-only model did the opposite. Specifically, our method achieves strong single-panel results while keeping almost the same multi-panel performance. 
This shows that our alternating training strategy helps the model retain knowledge from all levels, avoid the catastrophic forgetting problem, and handle the imbalance in data amount among coarse and fine levels.

\section{Conclusion and Discussion}
In this work, we introduced Panel2Patch, an automatic pipeline that converts multi-panel biomedical figures into large-scale hierarchical vision--language corpora with supervision at the figure, panel, and region levels, and showed that pretraining in this space yields consistent gains across retrieval, grounding, and zero-shot tasks. By explicitly modeling the nested figure--panel--region layout of scientific figures and using LVLMs for panel parsing, grounding, and region captioning, we turn existing literature into multi-scale supervision without additional human annotation, providing a cost-effective way to scale biomedical vision--language pretraining. Our experiments demonstrate that Panel2Patch improves both single-panel and bounding-box–text retrieval, strengthens fine-grained alignment, and, with alternating training, preserves multi-panel capability. More broadly, our results suggest that exploiting the compositional structure of scientific figures and mining high-quality hierarchical supervision can be as important as increasing dataset size for biomedical foundation models.

\pagebreak
\section{Acknowledgment}
This work benefited from French state funds managed by the National
Research Agency via the IHU Strasbourg (ANR-10-IAHU-02) and the ENACT AI
Cluster (ANR-23-IACL-0004). This work has received funding from the
European Union (ERC, CompSURG, 101088553). Views and opinions expressed
are however those of the authors only and do not necessarily reflect
those of the European Union or the European Research Council. Neither
the European Union nor the granting authority can be held responsible
for them. Serena Yeung is a Chan Zuckerberg Biohub — San Francisco Investigator.

\clearpage
\setcounter{page}{1}
\maketitlesupplementary

\section{Panel2Patch Details}
\label{sec:panel2patch}

Here we provide additional implementation and prompt-design details for our proposed {Panel2Patch} pipeline, which converts complex biomedical figures into a set of localized, semantically grounded patches. The pipeline consists of four main stages: (i) SoM-guided panel decomposition, (ii) marker-guided region mining, (iii) caption-box mining, and (iv) bounding-box post-processing. Unless otherwise stated, all stages use the same LVLM configuration described in Sec.~\ref{sec:lvml-configs}.

\subsection{Prompts for SoM-Guided Panel Decomposition}
\label{sec:prompt_panel_decomposition}

\begin{figure*}
    \centering
    \includegraphics[width=\linewidth]{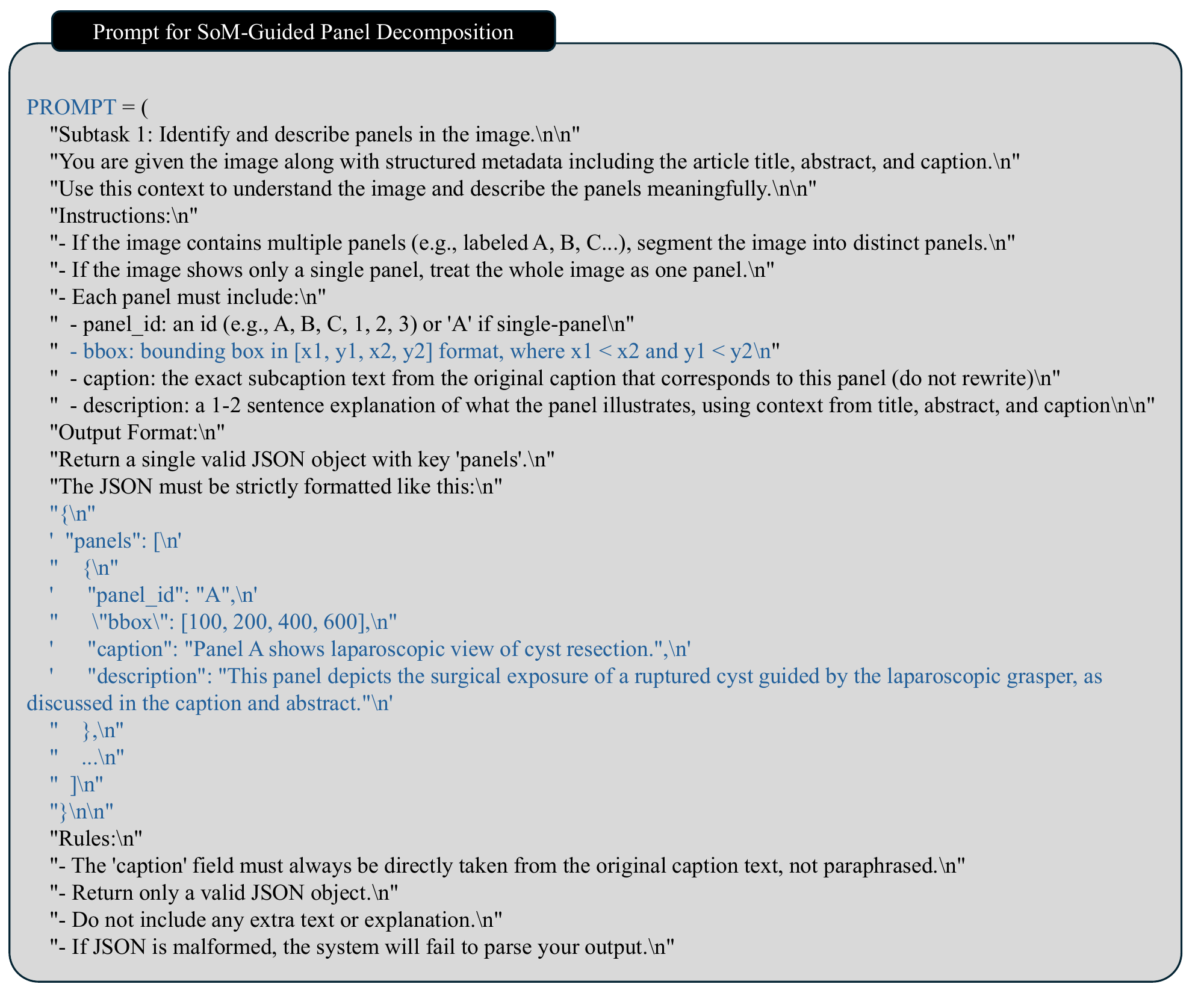}
    \caption{\textbf{Prompt for SoM-guided panel decomposition.} We show an example prompt and LVLM response for decomposing a multi-panel biomedical figure into individual panels, each with a panel ID, bounding box, and short description. The prompt enforces a strict JSON schema, which facilitates reliable downstream parsing.}
    \label{fig:prompt_panel_split}
\end{figure*}

As shown in Fig.~\ref{fig:prompt_panel_split}, we design a structured prompt that instructs the LVLM to first identify the global layout of a multi-panel figure and then enumerate all constituent panels together with their spatial extents. The prompt explicitly asks the model to:
\begin{enumerate}
    \item Determine whether the input is a single-panel or multi-panel figure.
    \item If multi-panel, list all panels in reading order (e.g., A, B, C, \dots).
    \item For each panel, return an axis-aligned bounding box in normalized coordinates $(x_{\min}, y_{\min}, x_{\max}, y_{\max})$ and a short description summarizing its content.
\end{enumerate}

To make parsing more robust, we embed SoM-style few-shot exemplars in the prompt that illustrate typical biomedical layouts, including grids of microscopy images, plots with shared axes, and schematic diagrams. These examples help the LVLM distinguish true content panels from auxiliary elements such as legends, color bars, or scale bars, which should not be treated as separate panels.

\subsection{Prompts for Marker-Guided Region Mining and Captioning}
\label{sec:prompt_marker_caption}

\begin{figure*}
    \centering
    \includegraphics[width=\linewidth]{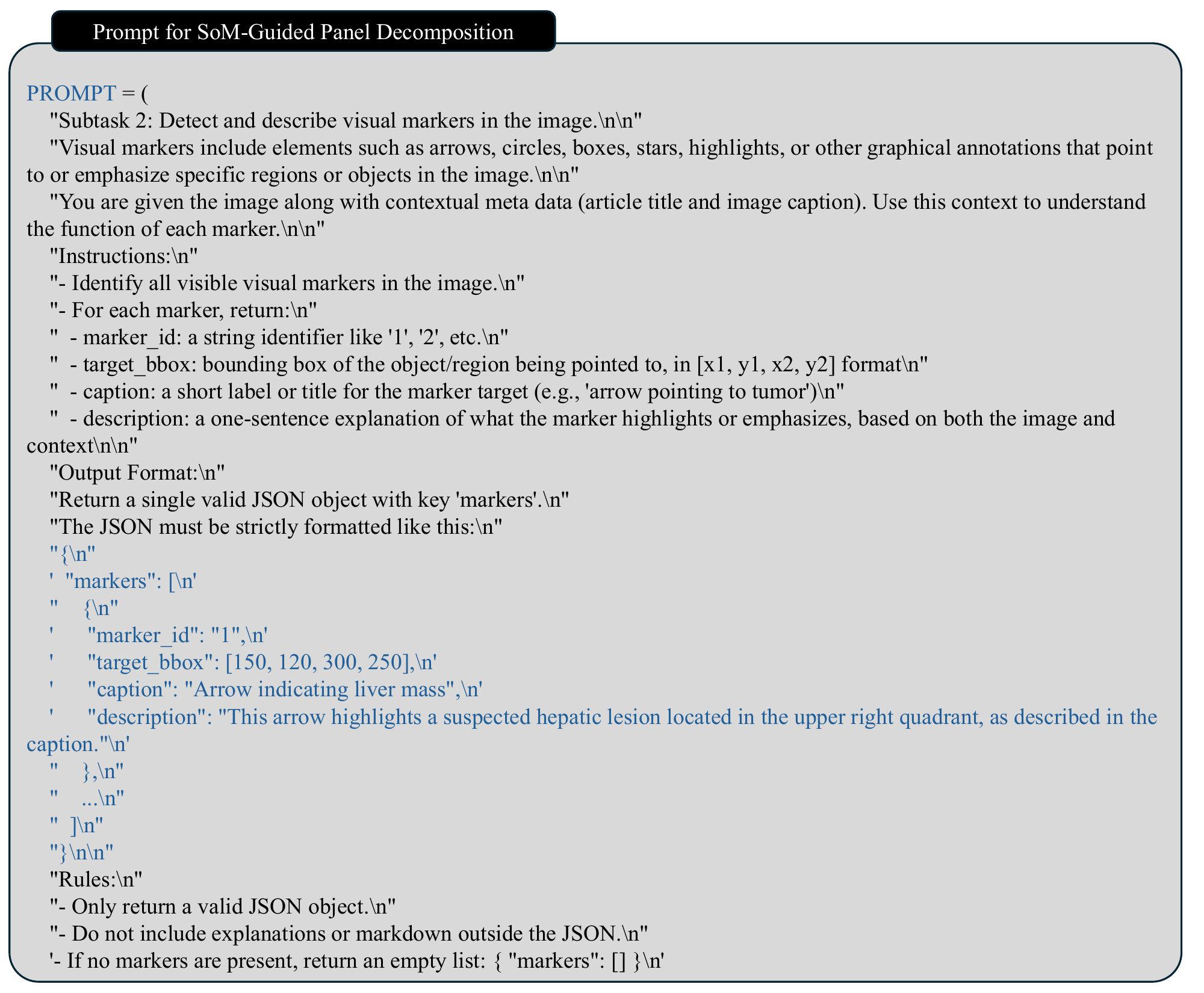}
    \caption{\textbf{Prompts for marker-guided region mining.} Example prompt and LVLM output for detecting visual markers (arrows, stars, etc.) and producing bounding boxes for the corresponding target regions. Each entry includes a marker ID, target box, short label, and a one-sentence description.}
    \label{fig:prompt_marker_bbox}
\end{figure*}

\paragraph{Marker Boxes.}
We further exploit fine-grained visual cues such as arrows, circles, stars, and other graphical markers that highlight regions of interest in biomedical figures (e.g., lesions, cells, or anatomical structures). As illustrated in Fig.~\ref{fig:prompt_marker_bbox}, we prompt the LVLM to:
\begin{enumerate}
    \item Identify all visible visual markers in the image.
    \item For each marker, infer its semantic role (e.g., ``lesion'', ``tumor boundary'', ``positive cells'') using the figure caption and article title as context.
    \item Return the bounding box of the \emph{target region} being highlighted, rather than the marker glyph itself.
\end{enumerate}
By grounding markers in the regions they point to, {Panel2Patch} captures subtle but clinically relevant structures that may occupy only a small fraction of the panel.

\begin{figure*}
    \centering
    \includegraphics[width=\linewidth]{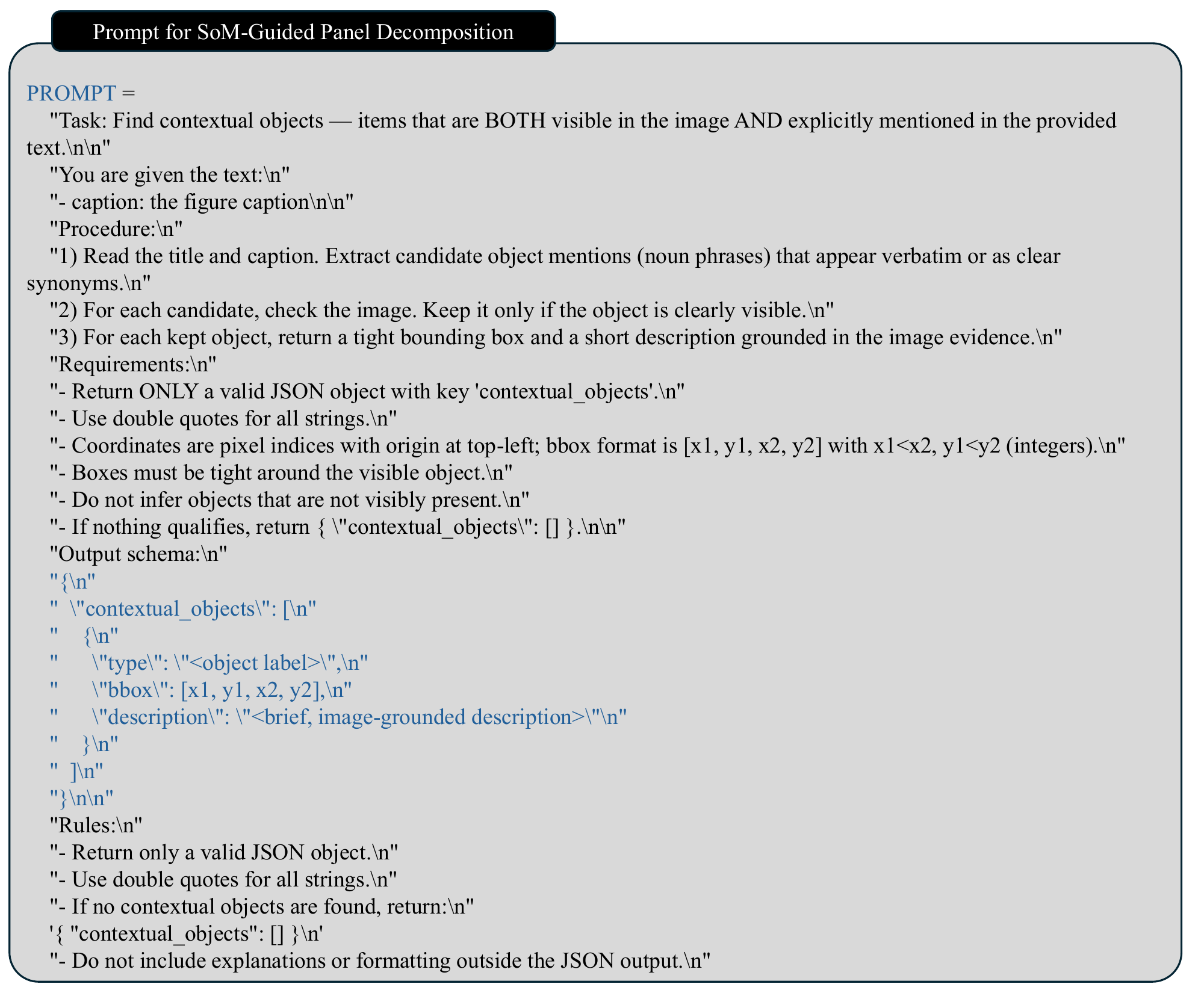}
    \caption{\textbf{Prompts for caption-guided region mining.} Example prompt and LVLM response for extracting bounding boxes and descriptions of contextual objects that are jointly grounded in the figure caption and the image content. Only objects that are explicitly mentioned and visually present are retained.}
    \label{fig:prompt_caption_bbox}
\end{figure*}

\paragraph{Caption Boxes.}
In addition to explicit markers, many figures contain local captions, labels, or inset annotations that implicitly define regions of interest (e.g., ``Zoomed-in view of region A'', ``Tumor core'', ``Control''). As shown in Fig.~\ref{fig:prompt_caption_bbox}, we design a complementary prompt that:
\begin{enumerate}
    \item Parses the caption text to extract candidate objects and regions explicitly mentioned in the figure description.
    \item Verifies that each candidate object is visibly present in the image.
    \item Returns tight bounding boxes and short descriptions for objects that are both mentioned in the caption and clearly visible.
\end{enumerate}
This caption-guided mining step focuses on ``contextual objects'' that are grounded in both text and image, complementing the marker-guided regions that are purely visually highlighted.

Combining marker-guided and caption-guided mining yields a richer set of semantically grounded regions, going beyond coarse panel crops to capture both visually emphasized and textually emphasized structures.

\subsection{LVLM Configs}
\label{sec:lvml-configs}

We run Qwen2.5-VL-72B in \textbf{float32} precision using vLLM with tensor parallel size $4$, GPU memory utilization $0.8$, maximum sequence length $8192$, and up to $512$ concurrent sequences. Decoding uses nucleus sampling with temperature $0.2$, top-p $0.9$, top-k $50$, repetition penalty $1.05$, and a maximum of $128$ generated tokens. Unless otherwise specified, these settings are used consistently for panel parsing, caption decomposition, marker-guided region mining, and region captioning.

\subsection{Bounding-Box Processing Details}
\label{sec:bbox_processing}

\begin{figure*}
    \centering
    \includegraphics[width=\linewidth]{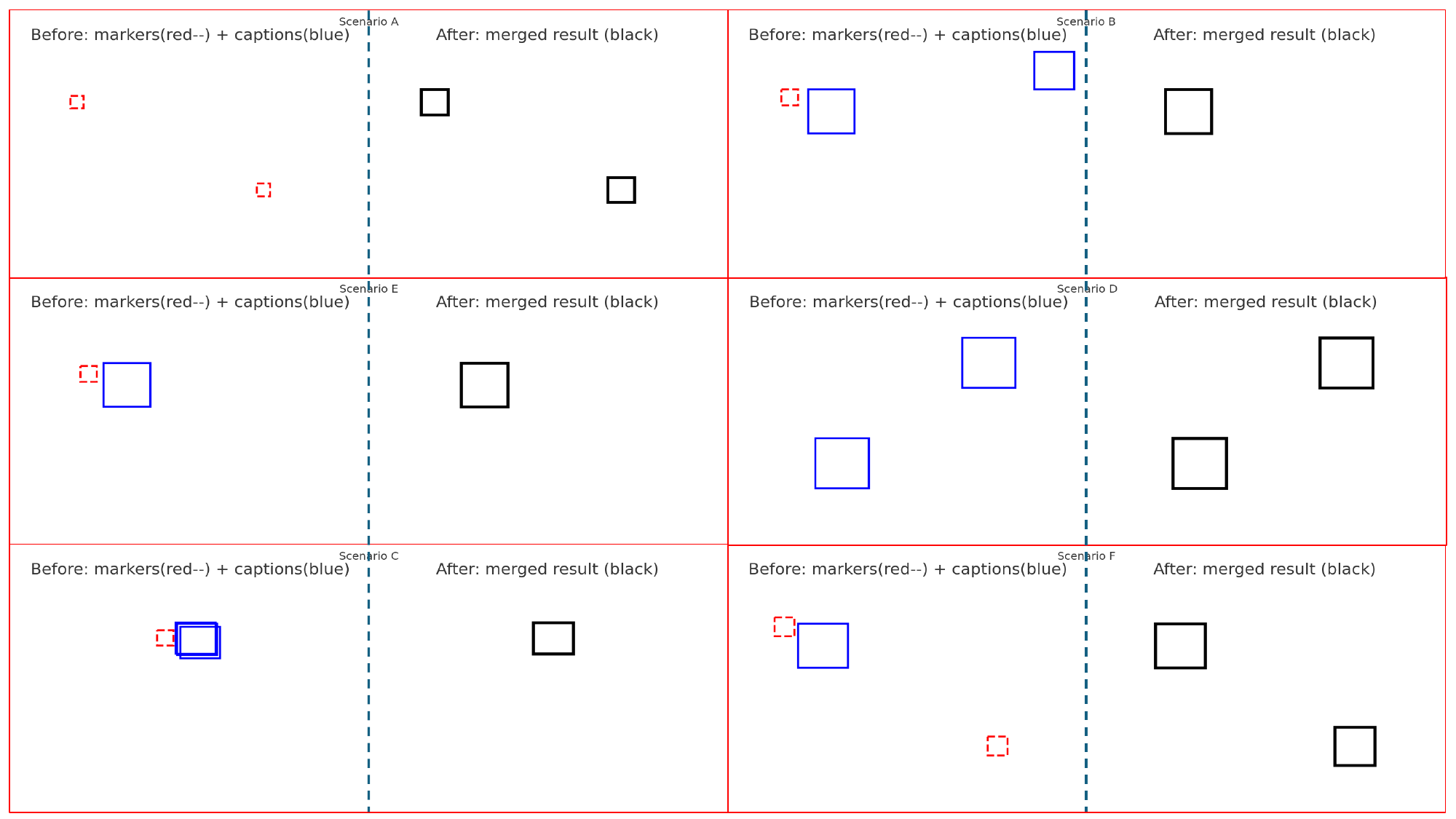}
    \caption{\textbf{Bounding-box post-processing pipeline.} Illustration of how our box post-processing method handles bounding boxes from different sources. We simulate different scenarios to demonstrate the method's robustness and precision.}
    \label{fig:bbox_process}
\end{figure*}

After obtaining raw bounding boxes from panel decomposition, marker mining, and caption mining, we perform several post-processing steps before constructing the final training set. Concretely, we:
\begin{enumerate}
    \item Convert all coordinates to absolute pixel values and clip them to the image boundaries.
    \item Discard degenerate boxes with extremely small area or aspect ratios outside a valid range.
    \item Apply class-agnostic non-maximum suppression (NMS) within each figure using an IoU threshold of $0.7$ to remove near-duplicate boxes produced by different prompts.
    \item Merge overlapping boxes that share highly similar textual descriptions, measured by a sentence-embedding cosine similarity above $0.9$.
    \item Normalize and store the final boxes as image--bbox--text triplets.
\end{enumerate}
This processing significantly improves the spatial precision and diversity of regions while avoiding excessive redundancy, as reflected by the visual examples in Fig.~\ref{fig:bbox_process} and Fig.~\ref{fig:bbox_process_visual}.

\section{Dataset Details}
\label{sec:data_details}

\subsection{Data Sampling}

We build our pretraining corpus on top of the Biomedica figure--caption dataset~\cite{lozano2025biomedicaopenbiomedicalimagecaption}. Starting from all available figures, we sample approximately $350\,\mathrm{k}$ multimodal figures and retain those that:
\begin{enumerate}
    \item provide an associated figure caption in English,
    \item are released under a license compatible with research usage,
    \item are annotated as scientific figures (excluding logos, advertisements, and non-scientific graphics),
    \item contribute to a reasonably balanced distribution of modalities.
\end{enumerate}
We then apply {Panel2Patch} to each sampled figure to obtain panel-level crops and region-level boxes, as detailed in Sec.~\ref{sec:dataset_stats}.

\subsection{Dataset Statistics}
\label{sec:dataset_stats}

Our final pretraining dataset contains:
\begin{itemize}
    \item $364{,}216$ figure-level image--caption pairs,
    \item $1{,}303{,}950$ panel-level image--caption pairs after panel decomposition,
    \item $619{,}424$ image--bbox--text triplets obtained from marker-guided region generation, and
    \item $1{,}030{,}194$ bounding boxes generated using the LVLM's internal knowledge (e.g., inferred object regions without explicit markers).
\end{itemize}
We take the union of marker-guided and LVLM-inferred boxes and apply the bounding-box processing procedure from Sec.~\ref{sec:bbox_processing} and Fig.~\ref{fig:bbox_process} before using them for pretraining.

The $364{,}216$ figure-level image--text pairs are drawn from Biomedica. We crawl the associated metadata and classify each figure into a primary and (optionally) secondary label corresponding to its dominant visual type (e.g., plots, microscopy, chemical structures). The distribution over secondary labels is long-tailed and reflects the diversity of real-world biomedical figures. As summarized in Tab.~\ref{tab:secondary_labels_part1} and Tab.~\ref{tab:secondary_labels_part2}, common categories include plots, bar plots, scientific illustrations, microscopy, and chemical diagrams, while many specialized categories (e.g., laryngoscopy, karyotype) appear at much lower frequencies. This diversity is beneficial for learning robust visual representations that generalize across modalities and scales.

\subsection{Quality of Annotations from Panel2Patch}
\label{sec:panel2patch_quality}

\begin{figure*}
    \centering
    \includegraphics[width=\linewidth]{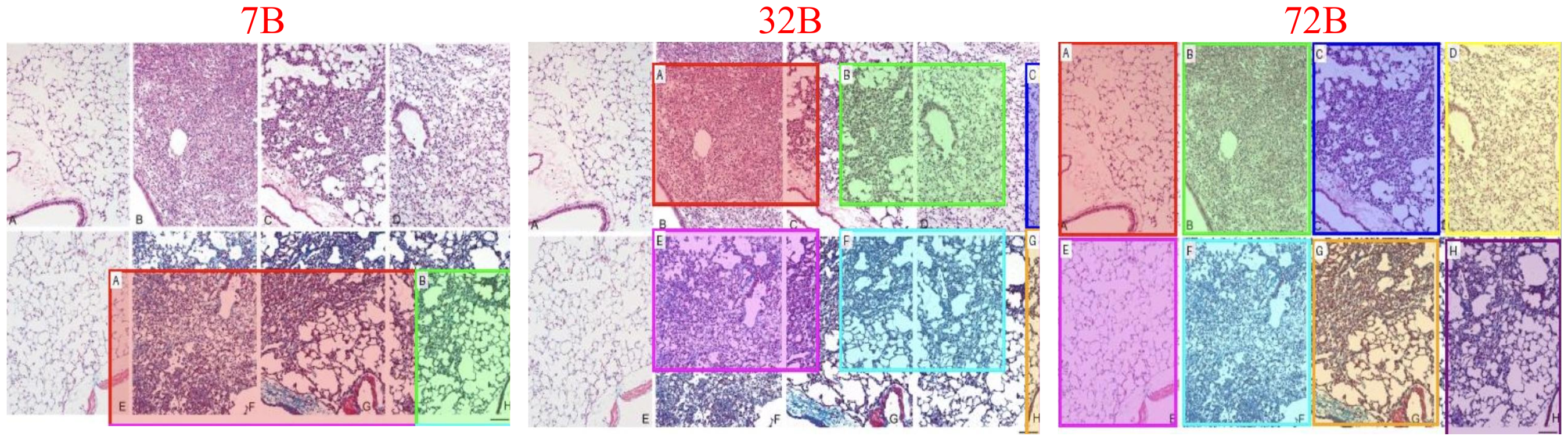}
    \caption{\textbf{Qualitative comparison of panel parsing.} Visualization of panel decomposition results for different LVLM capacities. From left to right, we show outputs from 7B, 32B, and 72B variants. Larger models (e.g., 72B) tend to produce more accurate panel boundaries and richer descriptions.}
    \label{fig:panel_process_visual}
\end{figure*}

\paragraph{LVLM for Panel Parsing.}
We empirically observe that stronger LVLMs lead to more accurate and consistent panel decompositions. As shown qualitatively in Fig.~\ref{fig:panel_process_visual}, Qwen2.5-VL-72B produces more precise panel boundaries, avoids merging unrelated subpanels, and yields more informative textual summaries compared to a smaller variant (e.g., 32B). In particular, the 72B model better respects subtle layout cues such as shared axes, legends, and insets, which are common in biomedical figures.

\begin{figure*}
    \centering
    \includegraphics[width=\linewidth]{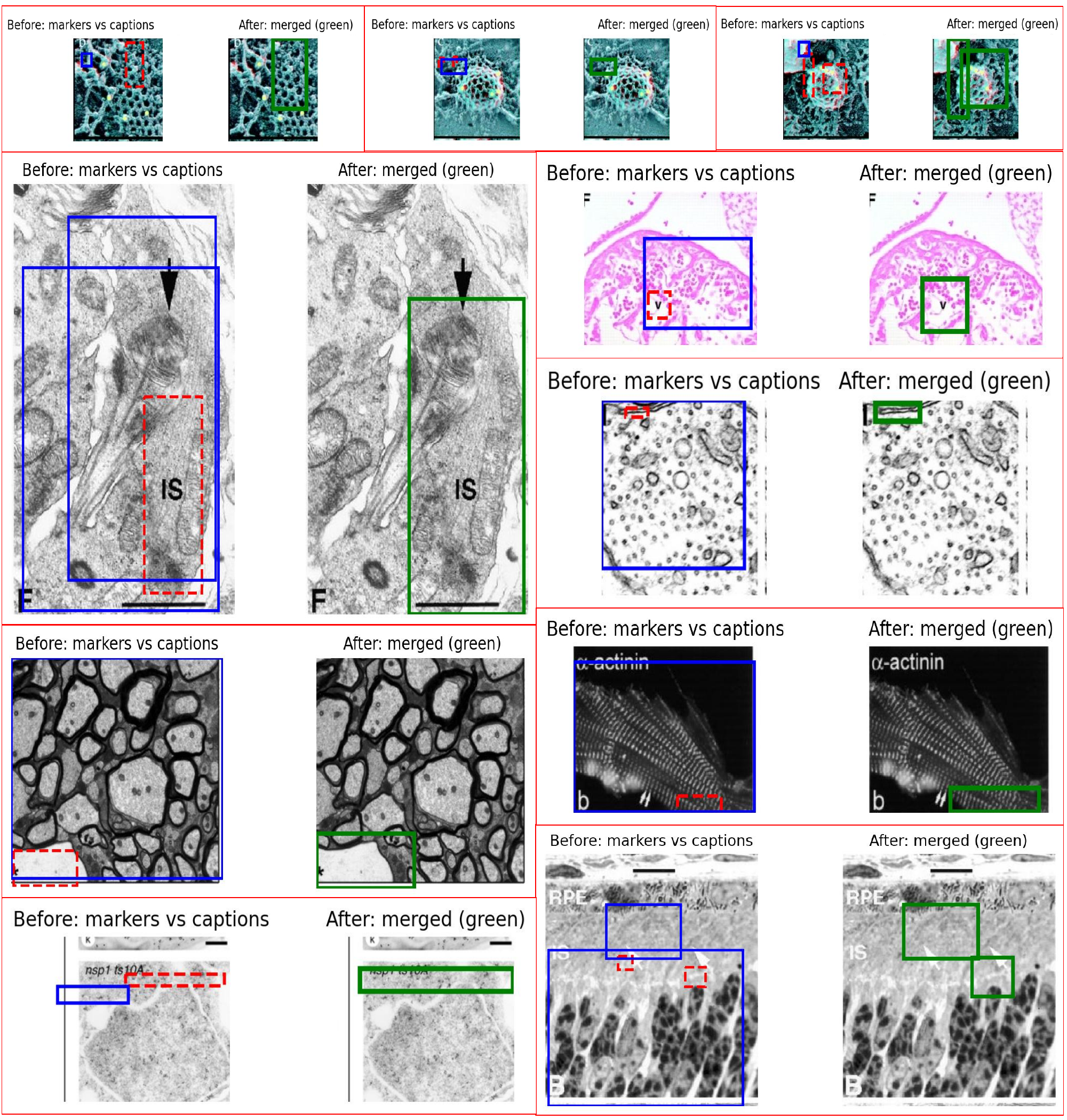}
    \caption{\textbf{Qualitative examples of region mining.} Visualizations of bounding boxes obtained from marker-guided and caption-guided mining after post-processing. Highlighted regions correspond to clinically or scientifically relevant structures emphasized in the figure.}
    \label{fig:bbox_process_visual}
\end{figure*}

\paragraph{LVLM for Biomedical Region Mining.}
For region-level mining, we find that LVLMs can reliably localize clinically meaningful structures when guided by explicit prompts about markers and captions. Fig.~\ref{fig:bbox_process_visual} shows qualitative examples where the model correctly highlights lesions, cell clusters, and anatomical regions indicated by arrows or local labels. Combined with the post-processing described in Sec.~\ref{sec:bbox_processing}, this yields high-quality region annotations suitable for training fine-grained biomedical vision--language models.

\begin{table*}[ht]
\caption{Secondary labels and percentages of figure types (Part 1).}
\centering
\begin{tabular}{ll|ll}
\hline
\textbf{Label} & \textbf{Pct} & \textbf{Label} & \textbf{Pct} \\
\hline
plot & 0.2392039961 & bar plot & 0.1150651605 \\
line plot & 0.0731292006 & scientific illustration & 0.0560098810 \\
2D chemical reaction & 0.0410442131 & microscopy & 0.0392800450 \\
diagram & 0.0284265115 & signal plot & 0.0271832018 \\
2D chemical structure & 0.0268386763 & light microscopy & 0.0218300065 \\
table & 0.0216904796 & immunohistochemistry & 0.0211481985 \\
flowchart & 0.0179900438 & immunoblot & 0.0178245154 \\
scatter plot & 0.0139932794 & matrix plot & 0.0127300424 \\
natural image & 0.0121686572 & forest plot & 0.0120581146 \\
map & 0.0120131838 & 3D chemical structure & 0.0106260247 \\
computerized tomography & 0.0106139191 & signaling pathway & 0.0103486139 \\
graph & 0.0095768316 & lab equipment & 0.0093916710 \\
fluorescence microscopy & 0.0093163344 & confocal microscopy & 0.0092534491 \\
3D plot & 0.0088056319 & electron microscopy & 0.0078804112 \\
box plot & 0.0074682040 & 3D protein structure & 0.0069536449 \\
heatmap plot & 0.0067048960 & clinical imaging & 0.0065234831 \\
laboratory specimen & 0.0055550288 & pie chart & 0.0053961494 \\
x-ray radiography & 0.0053488378 & assay & 0.0051444683 \\
specimen & 0.0045556989 & insects & 0.0040846959 \\
venn diagram & 0.0038361605 & cohort selection flowchart & 0.0037718538 \\
brain & 0.0033478943 & ambiguous & 0.0032435385 \\
skin lesion & 0.0030236729 & face & 0.0028432698 \\
ultrasound & 0.0028265426 & scanning electron microscopy & 0.0028222899 \\
surgical procedure & 0.0026714498 & histogram & 0.0026000397 \\
bacterial culture & 0.0023106903 & phylogenetic tree & 0.0021169777 \\
sequence plot & 0.0019832298 & nature & 0.0019563777 \\
endoscopy & 0.0018649276 & network & 0.0017417046 \\
electronic circuit & 0.0017399841 & epifluorescence microscopy & 0.0017307059 \\
tree & 0.0016858721 & angiography & 0.0016805242 \\
gel electrophoresis & 0.0016659331 & electrocardiography & 0.0016465339 \\
\hline
\end{tabular}
\label{tab:secondary_labels_part1}
\end{table*}

\begin{table*}[ht]
\caption{Secondary labels and percentages of figure types (Part 2).}
\centering
\begin{tabular}{ll|ll}
\hline
\textbf{Label} & \textbf{Pct} & \textbf{Label} & \textbf{Pct} \\
\hline
magnetic resonance & 0.0016456367 & dot plot & 0.0016198410 \\
functional magnetic resonance & 0.0016157048 & survival curve & 0.0014810793 \\
radial plot & 0.0014789354 & intraoral imaging & 0.0014599673 \\
human & 0.0014181240 & density plot & 0.0013346044 \\
circos plot & 0.0013059153 & user interface & 0.0011404918 \\
tool & 0.0010539856 & roc curve & 0.0009932869 \\
optical coherence tomography & 0.0009341144 & screenshot & 0.0008343880 \\
violin plot & 0.0008291993 & metabolic pathway & 0.0008278167 \\
phase contrast microscopy & 0.0008206435 & transmission electron microscopy & 0.0007905368 \\
aerial photography & 0.0007299856 & circular plot & 0.0007264203 \\
RT PCR & 0.0007226259 & teeth & 0.0006845110 \\
process chart & 0.0006718811 & checklist table & 0.0005735024 \\
3D chemical reaction & 0.0005326223 & intraoperative image & 0.0005291735 \\
procedural image & 0.0005042245 & neural network & 0.0004901149 \\
patient photo & 0.0004813765 & system diagram & 0.0004802114 \\
manuscript & 0.0004780481 & reagents & 0.0004636123 \\
humans and devices & 0.0004603306 & algorithm & 0.0004338552 \\
drawing & 0.0004153686 & immunoassay & 0.0003963423 \\
plot and chart & 0.0003894797 & flow diagram & 0.0003860271 \\
differential gene expression matrix & 0.0003814792 & medical equipment & 0.0003706203 \\
laryngoscopy & 0.0003408360 & 2D mesh & 0.0003278449 \\
eye & 0.0003246564 & flowcytometry & 0.0003243030 \\
human head & 0.0003202484 & word cloud & 0.0003137431 \\
skull & 0.0002876872 & list & 0.0002694608 \\
funnel plot & 0.0002647887 & immunocytochemistry & 0.0002596583 \\
illustration & 0.0002404416 & pyramid chart & 0.0001193004 \\
mammography & 0.0000844672 & karyotype & 0.0000812088 \\
\hline
\end{tabular}
\label{tab:secondary_labels_part2}
\end{table*}

\section{Implementation Details}
\label{sec:impl_details}

We perform continual pretraining by initializing from a strong Biomedica-trained CLIP backbone (ViT-L-14) and adapting it on the {Panel2Patch} corpus. Training is run with distributed data parallelism over $4$ GPUs using a per-GPU batch size of $40$ (effective batch size $160$) for $20$ epochs in \texttt{fp32} precision. We jointly optimize global figure-level objectives and region-level objectives for panels, fine regions, and coarse regions.

\subsection{Hyper-parameters}
\label{sec:hyperparams}

We use AdamW throughout with $\beta_1 = 0.9$, $\beta_2 = 0.95$, and weight decay $0.05$. Optimization is decoupled into a shared backbone branch and three region heads (bbox, fine, coarse), each with its own learning rate and cosine schedule:
\begin{itemize}
    \item shared/global branch: learning rate $1\times 10^{-5}$,
    \item bbox head: learning rate $5\times 10^{-6}$,
    \item fine head: learning rate $1\times 10^{-5}$,
    \item coarse head: learning rate $1\times 10^{-5}$.
\end{itemize}
All four branches use cosine learning-rate decay with a linear warmup of $1{,}000$ updates.

\subsection{Alternating Training}
\label{sec:alt_training}

To preserve general-purpose captioning ability and mitigate catastrophic forgetting, we alternate {Panel2Patch} data with the original Biomedica figure--caption pairs~\cite{lozano2025biomedicaopenbiomedicalimagecaption}. Concretely, we sample $1$ million figure-level image--caption pairs from the raw Biomedica dataset and mix them with our structured panel/region data. During training, each mini-batch is drawn from this mixture, and within {Panel2Patch} samples we randomly interleave bbox, fine, and coarse crops. This setup keeps the model exposed to both holistic figure-level captions and fine-grained region-level supervision, improving localized grounding while maintaining performance on generic figure captioning benchmarks.

\section{External Evaluation Benchmarks}
Our evaluation benchmarks in the main paper are strictly non-overlapping with the pretraining data.

\subsection{Cross-modal Retrieval}
As shown in Tab.~2 of the main paper, we consider two retrieval-based setups: (i) figure-level retrieval and (ii) region-level retrieval. To avoid contamination, we construct both benchmarks from PubMed articles published in 2025 that are not part of Biomedica~\cite{lozano2025biomedicaopenbiomedicalimagecaption} and are never used in our pretraining pipeline. For each benchmark, we run the same panel splitting and bounding-box generation pipeline used for pretraining. This results in two evaluation sets that mirror our pretraining data distribution while remaining strictly disjoint in both images and text. We additionally perform manual checks to remove any residual overlaps.

\subsection{Zero-shot Classification}
For zero-shot classification experiments, we follow the setup of~\cite{sun2025no} to ensure a fair comparison. In particular, we adopt the same label spaces, test splits, and evaluation metrics. Our method differs only in the visual--textual encoder and training data; we do not introduce any task-specific tuning or additional supervision beyond what is permitted in that protocol.

\subsection{Ablation Studies}
Due to the cost of continual pretraining, we conduct ablations on a reduced subset of the pretraining corpus. Specifically, we sample an additional $100\,\mathrm{k}$ figures from our processed dataset and re-run training under modified settings (e.g., removing marker-guided regions or LVLM-generated boxes). This allows us to isolate the contribution of each component while keeping training affordable.

\subsection{Cell Imaging}
\label{sec:cell_imaging}

We further evaluate our approach on cell imaging tasks to assess its ability to capture fine-grained morphological cues.

\begin{figure*}
    \centering
    \includegraphics[width=\linewidth]{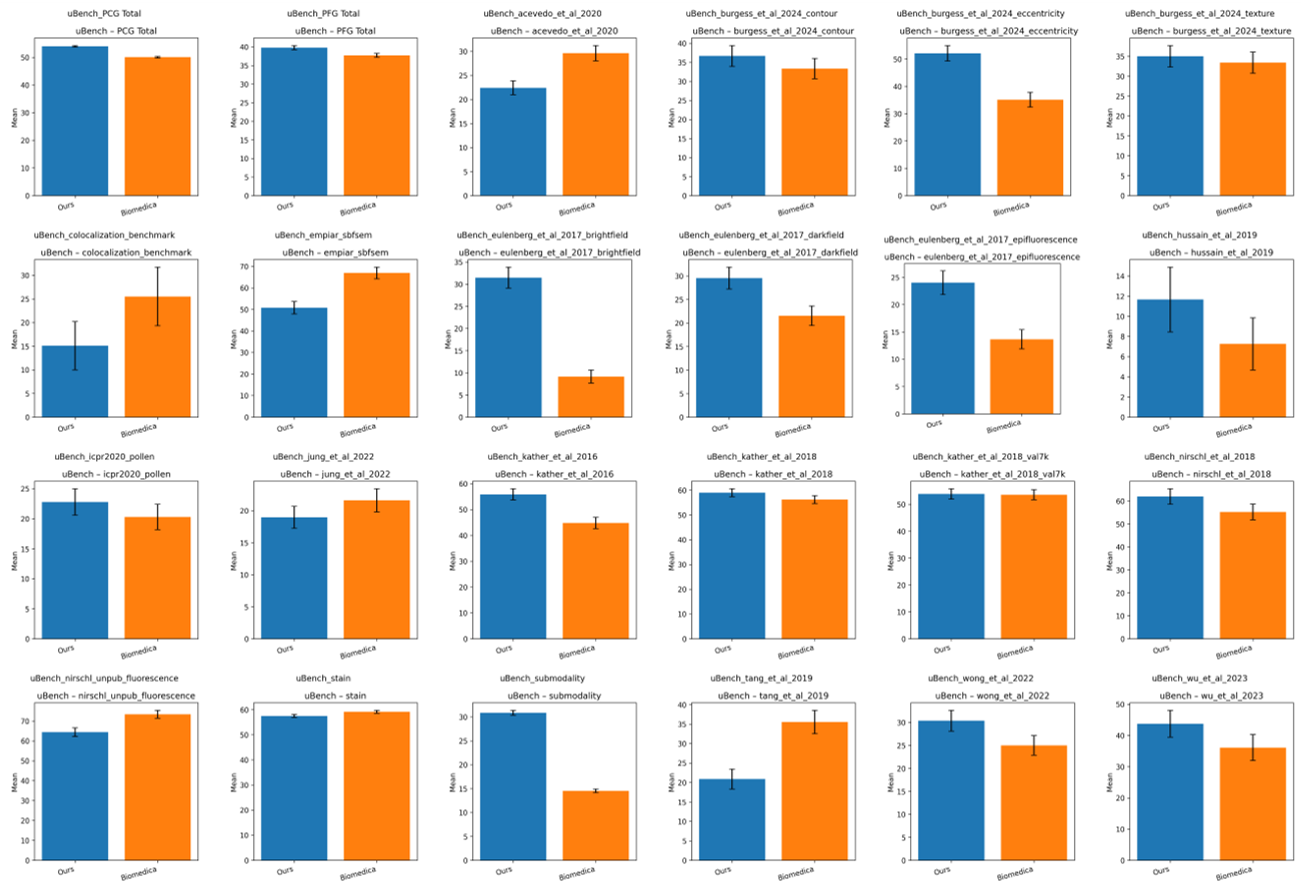}
    \caption{\textbf{Qualitative results on cell imaging.} Example retrieval and captioning results on a cell-imaging benchmark. The model trained with {Panel2Patch} produces more fine-grained and visually grounded descriptions compared to figure-level training, particularly for subtle morphological differences.}
    \label{fig:visualize}
\end{figure*}

As illustrated in Fig.~\ref{fig:visualize}, we visualize retrieval and captioning results on a challenging cell-imaging benchmark (e.g., fluorescence or bright-field microscopy). Our model is able to localize relevant subcellular regions (e.g., nuclei, cytoplasm, membrane structures) using the mined patches. It can also retrieve semantically similar cell images given a textual query describing morphological patterns (e.g., ``cells with fragmented nuclei'') and generate concise and accurate descriptions of cell states (e.g., apoptosis, mitosis, or abnormal morphology).

Qualitatively, the model trained with {Panel2Patch} better focuses on the discriminative parts of the cells than a baseline trained only on figure-level data, leading to improved alignment between textual descriptions and visual evidence.

{
    \small
    \bibliographystyle{ieeenat_fullname}
    \bibliography{main}
}

\end{document}